\documentclass[11pt,a4paper]{article}
\PassOptionsToPackage{hyphens}{url}\usepackage[hyperref]{acl2019}
\usepackage{times}
\usepackage{latexsym}
\usepackage{xcolor}
\usepackage{url}
\usepackage{breakurl}

\aclfinalcopy 

\title{Shall I Compare Thee to a Machine-Written Sonnet? An Approach to Algorithmic Sonnet Generation}

\author{\textbf{John Benhardt,$^{1\dagger}$ Peter Hase,$^{1,2,\dagger}$ Liuyi Zhu,$^{1,\dagger}$ Cynthia Rudin$^1$} \thanks{\ \ $\dagger$ denotes equal contribution.}\\
\textsuperscript{1}Duke University\\
\textsuperscript{2}UNC Chapel Hill\\
{\small \texttt{john.benhardt@duke.edu, peter@cs.unc.edu, liuyi.zhu@duke.edu, cynthia@cs.duke.edu}}
}

\begin{document}
\maketitle
\begin{abstract}
We provide an approach for generating beautiful poetry. Our sonnet-generation algorithm includes several novel elements that improve over the state of the art, leading to metrical, rhyming poetry with many human-like qualities. These novel elements include in-line punctuation, part of speech restrictions, and more appropriate training corpora. Our work is the winner of the 2018 PoetiX Literary Turing Test Award for computer-generated poetry. 
\end{abstract}

\section{Introduction}

Automated poetry generation is an extremely difficult task; currently no algorithm is able to capture the complexity of language well enough to generate truly meaningful poetry with all the nuances expected from a human poet \cite{deep-speare}. In the leading paradigm of poetry generation algorithms, which are founded in statistical learning of human poetry, humans must strike a fine balance between granting an algorithm expressive freedom and constraining it to follow poetic rules. This process requires finding good inductive biases for the learning of grammar, rhyme, meter, poetic style, and semantic content, as well as enforcing the sampling constraints needed to satisfy the relevant poetic rules. 

One particular form of poetry, the Shakespearean sonnet, provides a special challenge for algorithms. Each line in these poems must fit the stress pattern of iambic pentameter and the poem as a whole must follow the rhyming scheme of \textit{ABABCDCDEFEFGG}. These constraints, in addition to standard rules of grammar, must be obeyed when generating poems. 

In this paper we present a new approach for generating Shakespearean sonnets. By combining neural language models and expert rules, we demonstrate how certain inductive biases and sampling constraints can greatly improve poetry generation. Poetry generated by our algorithm is shown in Table \ref{lead}.\footnote{All examples from our approach shown in this paper are drawn from the same set of 50 generated sonnets. More examples are available in the supplement.}

\begin{table}

\framebox{\parbox{\dimexpr\linewidth-2\fboxsep-2\fboxrule}{

\vspace{2mm}
\textit{Quatrain and Couplet} -- Prompt: Sad

\medskip

\noindent The earth and of a woman to lament,\\
I see that you are in the eastern game\\
To me the odor of his hat and scent,\\
And if it shall be dear to me his name.\\
And now I say it is without surprise\\
It is the odor of his neck and eyes.

\vspace{3mm}
\hrule
\vspace{3mm}

\textit{Full Sonnet} -- Prompt: Love

\medskip

\noindent To show that it is any more than all,\\
And that the land is to ability\\
Than all that stretches, it begins to crawl\\
Of all I say it shall be dear to me.\\
Are not the song of what it is in doubt,\\
And I will feel the rush of heaven lined\\
And lessons of the sun and nights throughout\\
That it cannot be fittest for his mind.\\
The man that sprang it forth to glorify\\
I see the secret of the sun and light\\
I see just as good in the midst of my,\\
I see the workmen of the day and night.\\
Whoever you are in the eastern game, \\
And that it is the odor of his name.

\vspace{2mm}

}}

\caption{Samples generated by our approach.}

\vspace{-3mm}

\end{table} \label{lead}

Comparing our algorithm to past work, we show that poetry quality is noticeably improved by the addition of in-line punctuation, part-of-speech-based sampling constraints, and careful selection of language model training corpora. As poetry is ideally judged by humans, we discuss an extrinsic evaluation of our results.\footnote{The code for our work is available at \burl{https://github.com/peterbhase/poetry-generation}.}

\section{Related Work}

Poetry generation techniques rely on two general methods: rule-based expert systems and statistical learning methods. Expert systems have made use of grammatical, metrical, and rhyming templates, drawing words from source texts of varying sizes \cite{oliveira2009,oliveira2012,gervas2000,colton2012} or knowledge bases built from larger corpora \cite{netzer2009}. 

\citet{greene} provided an initial use case of a statistical language model, and since then neural language models have been used widely in production of English, Spanish, and Chinese poetry \cite{topical_poetry,hafez,rhythmic_verse,deep-speare,yi2018}. The core of these approaches is to train a neural language model on poetry and then sample from the model with constraints designed to satisfy poetic rules.


Our approach is most similar to those of \citet{topical_poetry} and \citet{deep-speare}; both of these approaches use constrained sampling from a neural language model to generate Shakespearean sonnets. 

Relative to the work of \citet{topical_poetry}, we improve poetry generation by means of adding in-line punctuation, adding part-of-speech-based grammatical constraints, and using training corpora more appropriate to the sonnet format.

Our work was performed concurrently with that of \citet{deep-speare}, and the conceptual differences between our approaches are two-fold: First, as we sample from the language model, we follow hard constraints on meter and rhyme, whereas they enforce soft constraints suggested by meter and rhyme models. Second, we select lines via a beam search to approximately optimize for line likelihood under the model, while they generate lines by sampling word by word from the language model using a low-temperature softmax. Consequently, our poetry follows the metrical and rhyme constraints of sonnets more precisely and each line will tend to have a higher likelihood than the lines in their sonnets, as computed by their respective language models.

\section{Approach}

The base of our approach is shared in common with the Hafez algorithm of \citet{topical_poetry}. At a high level, the scheme is to first pick the rhyme words and then use a word-level RNN to build each line backwards from the rhyme words.

\subsection{Base of Approach}

Critical to the Hafez approach was the Carnegie Mellon University Pronouncing Dictionary,\footnote{http://www.speech.cs.cmu.edu/cgi-bin/cmudict} which contains phoneme and stress patterns for over 134,000 words. Our approach uses two key pieces of information from the dictionary: the phoneme structure for each word and the stress pattern for each word. While the CMU dictionary includes three stress patterns, we simplify the dictionary to include only unstressed ($0$) and stressed ($1$) syllables. We use this dictionary to ensure that all generated lines conform to iambic pentameter and follow the necessary rhyme pattern. 
 
Generation begins with a user supplied prompt. The first step is to pick rhyme words. To do so, we compute a similarity score between the GloVe vector representation of the user-supplied topic and all metrically valid rhyme pairs \cite{pennington2014glove}. This similarity score is the maximum cosine similarity between the topic representation and each of the two words in the rhyme pair. If the user-supplied topic consists of several distinct tokens, we take the average of their word embeddings as the topic representation. We then transform the rhyme pair similarities into a probability distribution and sample rhyme pairs from the distribution, favoring rhyme pairs with high similarities. To conform with the sonnet rhyme scheme, each time a pair is drawn, all other pairs with that rhyme phoneme are discarded from the distribution and the probabilities are renormalized. 

Given the rhyme words, we begin constructing each of the fourteen lines of the sonnet independently. We use a word-level RNN with LSTM units to evaluate the likelihoods of metrically valid sequences \cite{lstm}.\footnote{Hyperparameters are given in the supplement.} In doing so, we employ a beam search, selecting for the highest likelihood paths. We sample each line from the 10 highest likelihood lines returned by the beam search. For search widths under 20 words, poems can be generated within 10 minutes on a standard laptop CPU. 

\subsection{Novel Elements in Generation}

Here we describe a few improvements over the previous state of the art represented in \citet{topical_poetry}, and we corroborate the usefulness of some techniques that appear in the concurrent work of \citet{deep-speare}. 

\subsubsection{Adding In-line Punctuation} 
We have found that in-line punctuation makes poems far more representative of actual human poetry, which often reads more slowly than prose due to comma placement. 

We add commas in a post-processing step, after all of the words for a poem have been chosen. Our approach is motivated by an issue inherent to selecting for punctuation during the construction of the poem. Punctuation does not affect the meter, which means that in order for a beam search to search over lines including punctuation, possible next words would always need to be considered with and without punctuation preceding them. This would increase the search cost by a factor of roughly $2^{d}$, where $d$ is the search depth. 

We circumvent this cost by approximately maximizing the line likelihoods, conditioned on the poem having some number $n$ commas. Once the words for a poem have been selected, our algorithm samples a number $n$ of commas to place into the poem. Then it evaluates the likelihood of a comma in every possible position, before placing commas into the $n$ most likely positions. We choose a distribution over the number of commas that produces a desirable amount of punctuation, in our judgment. This method only approximately optimizes the line likelihoods, since the ranking of the most likely positions could change after every comma placed into the poem. In practice, we observe that the approach tends to place commas in appropriate locations. 

Thus we achieve the presence of punctuation as \citet{deep-speare} do, but unlike in their work, we still optimize for line likelihood. 

\subsubsection{Part of Speech Restrictions}

We observe that across competitive approaches to poetry generation, part-of-speech (PoS) errors occur that could be easily detected by a human (see Section \ref{comparison}). Using the Python NLTK package's PoS tags, we reviewed poems generated by an earlier version of our program and recorded the tag sequences that produced ungrammatical phrases. We then used this knowledge to restrict the possible PoS paths in generated lines to ensure that the ungrammatical PoS sequences would not appear. For example, a pronoun typically does not directly precede a pronoun in English (e.g. ``he it"), so we add a restriction to prevent this from occurring. We created rules in this form for the 35 parts of speech that occur in the NLTK package.

\subsubsection{Suitable Corpora}

Relative to the work of \citet{topical_poetry}, which used a training corpus of contemporary song lyrics, we rely on corpora more suited to sonnet generation, namely {\it Endymion} by John Keats, collected works of Walt Whitman, and the {\it Hunger Games} trilogy by Suzanne Collins.\footnote{These texts are all publicly available.} We found that, compared to using a combination of texts from a wide variety of authors, using works from a single author provided a more coherent voice to our generated poems. Meanwhile, the texts have their individual fits to the meter and cadence that sonnets require. {\it Endymion} is written in iambic pentameter, the meter used in sonnets, while Walt Whitman's style is more traditionally poetic than the song lyric corpus. When generating a poem, we randomly pick a language model trained exclusively on one author's works.

\citet{deep-speare} use a corpus of sonnets. This is certainly suitable for the task at hand. However, without hard sampling constraints on meter and rhyme, their method is incapable of producing sonnets if they were to train on texts like the Hunger Games. Consequently, they cannot write sonnets in the voice of a prose writer, as our approach and that of \citet{topical_poetry} can.

\begin{table} 

\framebox{\parbox{\dimexpr\linewidth-2\fboxsep-2\fboxrule}{

\vspace{2mm}

Ours
\smallskip

\textit{Quatrain 1} -- Prompt: Nature

\smallskip

\noindent Of it, I might be able to explain \\
The same for both of us along the threat \\
With a minute and, crashing through the plain,\\
Comparing what, I have to want it yet.

\vspace{2mm}
\hrule
\vspace{3mm}

\textit{Quatrain 2} -- Prompt: Sea

\smallskip

\noindent I get into the horn and calories,\\
It would be able to prepare for gales,\\
I take a family in the merchant seas\\
Enough for me to reach the district trails.

\vspace{2mm}

}}

\caption{Excerpted quatrains from sonnets generated by our approach.}

\label{ourPoems}
\end{table} 

\section{Evaluation}

In this section we compare our poetry to that of \citet{topical_poetry}\footnote{Obtained through the interface provided by \citet{hafez}} and \citet{deep-speare}.\footnote{Poems were generated from the \texttt{LM}$^{**}$+\texttt{PM}+\texttt{RM} model.} We also discuss an extrinsic evaluation of our work.

\subsection{Poem Comparisons} \label{comparison}

In Tables \ref{ourPoems} and \ref{otherPoems}, we show quatrains from our approach as well as that of \citet{topical_poetry} and \citet{deep-speare}. Through our PoS sampling constraints, our poetry avoids grammatical errors seen in Quatrains 3 and 5. By prohibiting directly adjacent nouns, our approach avoids such phrases as ``human beings procreation'' in 3, and a constraint on adjacent comparative adjectives and verbs would prevent ``greater be'' in 5. 

Meanwhile, Quatrains 5 and 6 fails to adhere to an acceptable rhyme scheme. \citet{deep-speare} report an F1 score of .91 when evaluating against the CMU dictionary, but even a single mistake in the rhyme can give the poem away as being machine-generated. By directly constraining based on the CMU dictionary, our approach always avoids these kinds of mistakes.

\subsection{Extrinsic Evaluation}

At present, poems' meter and rhyme can be automatically evaluated, but qualities like theme, imagery, and other important poetic flourishes must still be evaluated by human readers. Since humans sometimes bend formal poetic rules to achieve certain literary effects \cite{deep-speare}, an overall assessment of a poem is best given by a human reader who can holistically evaluate its quality. 

Poetry from our algorithm was assessed by a team of trained readers organized by Dartmouth's Neukom Insitute for the purpose of judging the 2018 PoetiX Literary Turing Test. In Turing test fashion, sonnets generated by our approach were mixed with human-generated sonnets and presented to the judges for them to distinguish between. The judges were able to successfully separate the human poems from ours, with a majority of judges labeling every case correctly. Yet while our poems failed to fool the judges, they did receive first place in the competition.

It is important to note that these judges were trained readers of sonnets, some with backgrounds in computer science and some in the humanities. \citet{deep-speare} find that ``meter is largely ignored by lay persons in poetry evaluation,'' and so a thorough evaluation requires expert readership. 

Our method succeeds that of \citet{topical_poetry} as a winner of the Neukom Institute's competition.

\begin{table} 

\framebox{\parbox{\dimexpr\linewidth-2\fboxsep-2\fboxrule}{

\vspace{2mm}
\citet{topical_poetry}
\smallskip

\textit{Quatrain 3} -- Prompt: Nature

\smallskip

\noindent That solar system needs a great salvation. \\
This world has never seen another kind, \\
Life as human beings procreation! \\ 
We know the greatest only humankind. 

\smallskip

\textit{Quatrain 4} -- Prompt: Love

\smallskip

\noindent I really wanna feel a little sadder! \\
I never wanna see another mother! \\
An old spectacular spectacular, \\ 
Or something like a letter from her lover. 

\vspace{2mm}
\hrule
\vspace{3mm}

\citet{deep-speare}
\smallskip

\textit{Quatrain 5}

\smallskip

\noindent and thank him, what he offers you in me \\
and i will only what a greater be \\ 
put on the doctors of his endless doom \\
run this, to him who left his ways to roam

\smallskip

\textit{Quatrain 6}

\smallskip

\noindent   but much then it is an greater man\\
so such as it is by day's seeing men\\
then, with a giddy motion of the sun\\
and to be cast of the desert, and crown
\vspace{2mm}

}}

\caption{Two poetry samples from each of \citet{topical_poetry} and \citet{deep-speare}. Note that the poems from \citet{deep-speare} are generated to follow any of the \textit{ABAB, ABBA}, and \textit{AABB} rhyme schemes.}

\label{otherPoems}
\end{table} 

\section{Conclusion}

We present an approach for algorithmic sonnet generation that capitalizes on inductive biases and expert rules to produce sonnets with many human-like qualities. While our poems are generated by approximate likelihood maximization, they also contain punctuation, always follow the metrical and rhyme-related rules of sonnets, and reflect the voice of the writer whose works they were trained on. Our approach was evaluated by expert readers in the Neukom Institute's 2018 Literary Turing Test, where it received first prize.

\nocite{endymion,whitman,hungerGames1,hungerGames2,hungerGames3}

\bibliography{acl2019}
\bibliographystyle{acl_natbib}

\end{document}


\maketitle

\begin{table}
	\centering
	\begin{tabular}{ll}
		\toprule
		Hyperparameter & {Value or Function}  \\
		\midrule
		Num. Layers & $3$ \\
		Layer Width & $1000$ \\
		Learning Rate Decay & \texttt{\string tf.train.cosine\_decay\_restarts} \\
 		\quad \texttt{learning\_rate} & $.0002$ \\
 		\quad \texttt{first\_decay\_steps} & $2e4$ \\
        \quad \texttt{t\_mul} & $1$ \\
        \quad \texttt{m\_mul} & $.1$ \\
        Batch Size  & $50$ \\
        Dropout & $.3$ \\
        Embedding Dim. & $300$ \\
        Epochs & $20$ \\ 
        Epoch When Embeddings Unfrozen & $15$ \\
		\bottomrule
	\end{tabular}
	\vspace{3mm}
	\caption{Hyperparameters for language model, which is an RNN with LSTM cells.}
	\label{table:epochs}
\end{table}

\begin{table}

\vspace{3mm}
\hrule
\vspace{3mm}

\noindent  The earth and of a woman to lament,\\
I see that you are in the eastern game\\
To me the odor of his hat and scent,\\
And if it shall be dear to me his name.

\vspace{3mm}
\hrule
\vspace{3mm}

\noindent I get into the horn and calories,\\
It would be able to prepare for gales,\\
I take a family in the merchant seas\\
Enough for me to reach the district trails.

\vspace{3mm}
\hrule
\vspace{3mm}

\noindent Together in a farm, and take it planned \\
To stop, and raise his finger in the game \\
That I just had to stop, and understand \\
And see if you can think of our name.

\vspace{3mm}
\hrule
\vspace{3mm}

\noindent To show that it is any more than all,\\
And that the land is to ability\\
Than all that stretches, it begins to crawl\\
Of all I say it shall be dear to me.

\vspace{3mm}
\hrule
\vspace{3mm}

\noindent Its cold for such an hour of protect\\
To tell them Paul was glad it came his wife\\
Within a season, or just the subject\\
It always seems to me for lack of life.

\vspace{3mm}
\hrule
\vspace{3mm}

\noindent Of it, I might be able to explain \\
The same for both of us along the threat \\
With a minute and, crashing through the plain,\\
Comparing what, I have to want it yet.

\vspace{3mm}
\hrule
\vspace{3mm}

\noindent In caves across the brook, releases sand\\
The cold before I blamed it on the game\\
Beyond the shelves of floor it will be strand\\
To stop it with a figure of the same.

\vspace{3mm}
\hrule
\vspace{3mm}

\noindent Because I tried to pick it in the coast\\
The rest of the explosion, and his heel\\
And try to haul him back into the roast,\\
I tried to haul him back into the seal.

\vspace{3mm}
\hrule
\vspace{3mm}

\caption{Excerpts from sonnets generated by our approach. All excerpts pulled from the same set of 50 generated sonnets.}

\end{table}